\documentclass[letterpaper, 10 pt, conference]{ieeeconf}
\IEEEoverridecommandlockouts

\usepackage{bm}
\usepackage{amsmath} 
\usepackage{amssymb}  
\usepackage{amsfonts}

\usepackage{subfigure}
\usepackage{graphicx}
\usepackage{hyperref}

\usepackage{booktabs}
\usepackage{multirow}
\usepackage{tablefootnote}
\usepackage{threeparttable}

\def\calP{\mathcal{P}}

\def\calL{\mathcal{L}}

\def\Rbb{\mathbb{R}}
\def\Rbb3{\mathbb{R}^3}

\title{\LARGE \bf
Outram: One-shot Global Localization via Triangulated Scene Graph and Global Outlier Pruning
}

\author{Pengyu Yin$^{1}$, Haozhi Cao$^{1}$, Thien-Minh Nguyen$^{1}$, Shenghai Yuan$^{1}$, Shuyang Zhang$^{2}$, Kangcheng Liu$^{1}$, \\ and Lihua Xie$^{1}$, \emph{Fellow, IEEE}
\thanks{$^{1}$Authors are with the Centre for Advanced Robotics Technology Innovation (CARTIN),  School of Electrical and Electronic Engineering, Nanyang Technological University, Singapore.
{\tt\small pengyu001, haozhi002, thienminh.nguyen, shyuan, kangcheng.liu, elhxie@ntu.edu.sg}}
\thanks{$^{2}$Authors are with the Department of Electronic and Computer Engineering, the Hong Kong University of Science and Technology, Clear Water Bay, Kowloon, Hong Kong SAR, China.
{\tt\small szhangcy@connect.ust.hk}}
\thanks{This research is supported by the National Research Foundation, Singapore under its Medium Sized Center for Advanced Robotics Technology Innovation.}
}

\begin{document}

\maketitle
\thispagestyle{empty}
\pagestyle{empty}

\begin{abstract}
One-shot LiDAR localization refers to the ability to estimate the robot pose from one single point cloud, which yields significant advantages in initialization and relocalization processes. 
In the point cloud domain, the topic has been extensively studied as a global descriptor retrieval (i.e., loop closure detection) and pose refinement (i.e., point cloud registration) problem both in isolation or combined. However, few have explicitly considered the relationship between candidate retrieval and correspondence generation in pose estimation, leaving them brittle to substructure ambiguities. 
To this end, we propose a hierarchical one-shot localization algorithm called Outram that leverages substructures of 3D scene graphs for locally consistent correspondence searching and global substructure-wise outlier pruning. Such a hierarchical process couples the feature retrieval and the correspondence extraction to resolve the substructure ambiguities by conducting a local-to-global consistency refinement. We demonstrate the capability of Outram in a variety of scenarios in multiple large-scale outdoor datasets. Our implementation is open-sourced: \url{https://github.com/Pamphlett/Outram}.
\end{abstract}

\section{Introduction} \label{intro}
LiDAR-based localization problems can be stated in the following general form. We are given a point cloud $\mathcal{P}$ produced by a LiDAR, and a reference point cloud $Q$, which can be either another frame of LiDAR scan \cite{zhang2014loam}, an accumulated submap \cite{chen2022direct}, or even the entire mapping space \cite{nguyen2023slict}. Given a point correspondence $i\in \mathcal{I}$, $\mathbf{p}_i \in \mathcal{P}$ and $\mathbf{q}_i \in \mathcal{Q}$ can be associated and represented in the residual function $r_i:=\|\mathbf{T}\mathbf{p}_i - \mathbf{q}_i\|\rightarrow \left[0,\infty \right)$ where $\mathbf{T}$ is the ground truth localization result we are searching for. 

While the estimation problem can be relatively easy in special cases (e.g., the size of point clouds is constrained or an approximation $\mathbf{T}_{initial}$ is known a priori), it can be hard in general \cite{lim2022single, yin2023segregator} due to limited descriptiveness of local features and computational complexity. These general, prior-free cases are what we encountered in relocalization or global localization problems. To address this problem, prior work \cite{kim2021scan, yuan2023std, cattaneo2022lcdnet, kong2020semantic, li2021ssc} usually break it down into a retrieval phase and a pose estimation phase, where several candidates are generated first and verified later for final pose estimation. 
These decoupled approaches, however, suffer from local ambiguities where the substructures in different LiDAR scans are similar, leading to false candidate retrieval. 


Rather than attempting to find the most appearance-similar keyframe through one single retrieval, we propose an algorithm called Outram for one-shot, accurate, and efficient global localization directly against a reference map. Different from existing works, we rely on local substructures of a 3D scene graph to generate locally consistent correspondences directly with the map, and further find the inlier correspondence that globally unifies substructures-wise consistency. This leads to the proposed local-to-global, hierarchical global localization algorithm that has the following contributions:
\begin{itemize}
    \item We propose a novel representation encoding substructure of 3D scene graphs for efficient locally consistent correspondence generation.
    \item Together with a subsequent graph-theoretic pruning module, we propose an accurate, efficient, and one-shot global localization pipeline for large-scale outdoor environments.
    \item Extensive experiments are conducted on publicly available datasets on a global localization setup, showing superior robustness compared with current state-of-the-arts. We further open-sourced our implementation to benefit the community.
\end{itemize}

\begin{figure}[t]
\centering
\includegraphics[width=8.2cm]{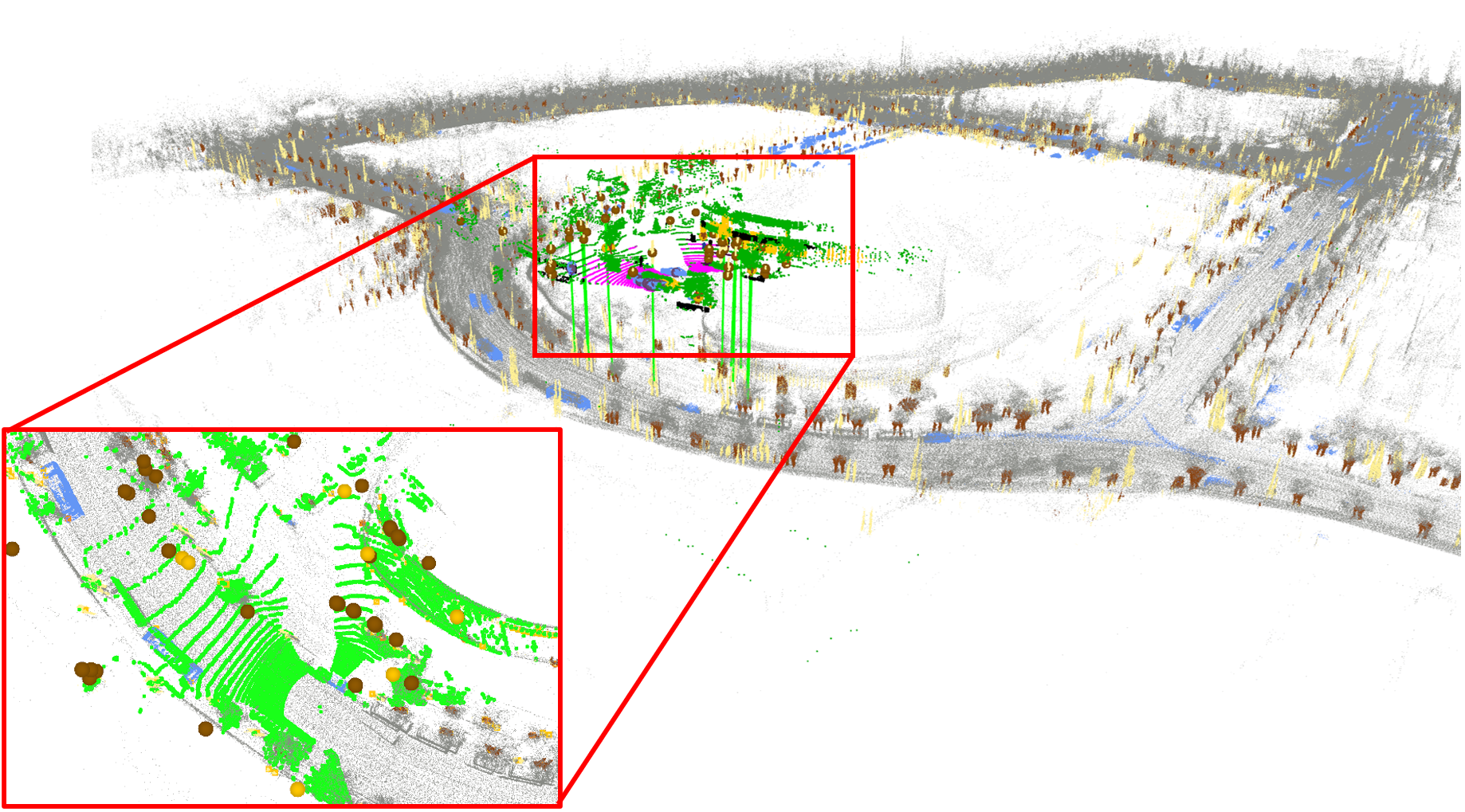}
\caption{Illustration of the proposed global localization algorithm on MulRan DCC dataset \cite{kim2020mulran}. We generate 3D scene graphs and leverage their substructures for locally consistent correspondence generation. With these raw correspondences, we exploit a graph-theoretic outlier pruning process for globally consistent inlier extraction (green lines). The point cloud transformed by the estimated pose is shown in green in the enlarged area. The position of the semantic segmented point cloud and the grey map point cloud is presented for visualization purposes only.}
\label{fig: openning}
\vspace{-1.2em}
\end{figure}

\section{Related Work}
In the literature, LiDAR-based relocalization or global localization methods can be broadly categorized into two branches by whether a movement of the robot is needed, namely one-shot localization and iterative localization. Iterative localization methods are usually formulated in a Monte Carlo localization manner \cite{kuang2023ir, wiesmann2023locndf}, where the movement of the robot will provide more environment observations and thus update the weight of each particle till convergence. With the accumulative submap and generated dense segments, Segmap \cite{dube2020segmap} proposed a data-driven retrieval mechanism for global localization. 

On the contrary, one-shot localization is referred to as algorithms that solve the global localization problem in a full prior-free manner. We further divide the literature on this into two groups: loop closure detection-based methods and registration-based methods. The next subsection provides more details on each of the two groups.

\subsection{Loop closure detection-based global localization}
Loop closure detection (LCD) methods identify previously visited places by encoding current LiDAR measurements into global descriptors and comparing them to a database constructed from historical frames. The construction process can be divided into global and local ones. Very similar to its visual-based counterpart, several local methods \cite{zhong2009intrinsic, steder2011place} detect 3D key points and aggregate them into a global representation, and the retrieval process is arranged in a bag-of-words manner. Alternatively, global methods directly encode the whole LiDAR scan. The encoding pattern can be either handcrafted or deep-learned. Scan Context \cite{kim2021scan} encodes the geometric information of a point cloud into a bird-eye-view global descriptor. Following the same data structure, several variants \cite{wang2020intensity, li2021ssc} have been proposed to enhance the descriptiveness by extending the original geometric-only representation by either the inherent intensity information \cite{wang2020intensity} or high-level semantics \cite{li2021ssc}. 
Several methods leverage deep neural networks to generate global descriptors directly \cite{uy2018pointnetvlad, chen2022overlapnet}. Uy et al. \cite{uy2018pointnetvlad} combined PointNet \cite{qi2017pointnet} and NetVLAD \cite{uy2018pointnetvlad} to generate compact global representations. Chen et al. \cite{chen2022overlapnet} proposed an LCD network that estimates the overlap ratio between point clouds. More recently, techniques exploiting the graph structure of local features have been proposed \cite{zhu2020gosmatch, kong2020semantic, pramatarov2022boxgraph, yuan2023std}. Methods involving point cloud semantics \cite{zhu2020gosmatch, kong2020semantic, pramatarov2022boxgraph} usually encode the scene by instances and the spatial layout in between. LCD is accomplished by conducting similarity checks between these semantic instance graphs. Yuan et al. \cite{yuan2023std} proposed to aggregate local point features to form a triangle-based descriptor (the simplest graph). Leveraging the side length of each triangle as the hash table key, loop closure candidates can be found by a voting scheme. Further, the transformation in between is calculated and verified by planes in the scene.

It is natural to extend LCD methods to global relocalization due to the similar retrieval mechanism \cite{xu2022lidar}. Nevertheless, local ambiguities and scene changes can make the descriptor-based algorithms brittle. Moreover, while geometric verification is widely employed after the retrieval process, it can be both time-consuming and inaccurate \cite{vidanapathirana2023spectral}.
Additionally, even if the geometrically proximate candidate is retrieved from the database, LCD methods usually rely on local point features for the subsequent pose estimation, which could potentially suffer from feature degeneracy \cite{lim2022single, yin2023segregator}, resulting in impossible pose estimation. 

\subsection{Registration-based methods}
Instead of finding one single keyframe in the pre-built LCD database, registration-based methods seek to solve the global localization problem in a point cloud registration manner. 
There exist two concerns \cite{pomerleau2015review} when leveraging point cloud registration in solving global localization problems: correspondence generation and outlier pruning. In correspondence extraction, it is not computationally feasible to directly extract correspondences on the point level. Subsequently, several methods leverage high-level representation in replace of geometric points. Ankenbauer et al. proposed \cite{ankenbauer2023global} semantic object maps for global localization by formulating a global registration problem, where all-to-all correspondence is built between semantic objects within local and global maps. In a very similar sense, all-to-all correspondence is also employed in \cite{matsuzaki2023single} for global localization, while semantic clusters to be registered are from different modalities. 

In the outlier correspondence pruning stage, the aforementioned two methods both send prebuilt correspondences into a graph-theoretic inlier selection module, where the inlier set is modeled as the maximum clique (MC) of the consistency graph \cite{yang2020teaser}.
Additionally, RANSAC-based methods \cite{cattaneo2022lcdnet} are also ubiquitous in the outlier pruning stage, while being proven to be brittle to high amounts of outliers \cite{yang2020teaser, shi2021robin}.

In comparison with these methods, we proposed a novel substructure representation, instead of semantic clusters only, of a semantic segmented point cloud for more informative correspondence extraction. We emperially demonstrate in \ref{sec: results} that how our proposition is more computationally tractable and accurate than previous state-of-the-art.

\section{Methodology}
In this section, we first formulate the global localization problem that is considered herein. Next, we present our proposed one-shot global localization algorithm Outram with two sub-modules, where we first leverage local substructures in a 3D scene graph for correspondence generation and further prune the correspondence globally with substructure-wise consistency check.

\subsection{Registration-based One-shot Global Localization}
The point cloud registration problem is formulated as acquiring the pose transformation $\mathbf{T}$ of a single query LiDAR point cloud $\mathcal{P}=\left\{\mathbf{p}_i\in\mathbb{R}^3\right\}_{i=1}^{n}$ against a prebuilt point cloud map $\mathcal{M}=\left\{\mathbf{m}_j\in\mathbb{R}^3\right\}_{j=1}^{m}$ accumulated by a series of scans in the world frame collected during a time span $\left[1, t\right]$. The pose transformation in between is defined as:
\begin{equation}
    \mathbf{T} \triangleq \left[\mathbf{R},\mathbf{t}\right] \in \text{SO(3)} \times \mathbb{R}^3,
\end{equation}
where $\mathbf{R}$ represents the rotation and $\mathbf{t}$ is the translation. With the unknown ground truth transformation, corresponding points in the query scan and the reference map can be associated as:
\begin{equation}
    \mathbf{m}_j=\mathbf{R} \mathbf{p}_i + \mathbf{t} + \mathbf{o}_{ij},
    \label{eq: generative func}
\end{equation}
where $\mathbf{o}_{ij}$ is the measurement error. Finding the pose transformation typically includes three steps: find an initial data association $\mathcal{I}\in[n]\times[m]:=\left\{1,\dots,n\right\}\times\left\{1,\dots,m\right\}$, prune the initial correspondences set $\mathcal{I}$ for the inlier set $\mathcal{I}^{\star}$, and estimate the pose transformation with the inlier set $\mathcal{I}^{\star}$ \cite{pomerleau2013comparing}. In the following sections, we will detail how we leverage scene graphs for informative correspondence generation and efficient outlier pruning. We also empirically demonstrated in \ref{sec: results} how our proposed method is superior to current existing works \cite{ankenbauer2023global} leveraging all-to-all correspondence for global localization problems in terms of scalability.


\subsection{Triangulated 3D Scene Graph} \label{Tri Scene Graph}
Since point feature level correspondence generation is not computationally feasible in global localization problems, we leverage 3D scene graphs for correspondence extraction at a higher instance level. Different from previous works \cite{ankenbauer2023global, zhang2023instaloc} that build all-to-all correspondence or leverage instance-only descriptor, we present a new representation, triangulated 3D scene graphs, for informative and efficient correspondence generation.

Given the query point cloud $\mathcal{P}=\left\{\mathbf{p}_i\right\}_{i=1}^{n}$, we employ the state-of-the-art point cloud semantic segmentation network \cite{cao2023multi} to match a point $\mathbf{p}_i$ with a semantic label $l \in \calL$. The network thus acts as a mapping $\lambda(\mathbf{p}_i) \colon \mathbb{R}^3 \to \mathbb{\calL} \subset \mathbb{N}$.
Hence, we can define the following semantic $\mathcal{S}$ point cloud as:
\begin{equation}
    \mathcal{S}=\left\{s_i | s_i=\left(\mathbf{p}_i,\lambda\left(\mathbf{p}_i\right)\right),\forall\mathbf{p}_i\in\mathcal{P}\right\}.
\end{equation}

Subsequently, we leverage the projection-based clustering method  \cite{zhou2021t} to generate instances from the semantic point cloud with the same label $l$:
\begin{equation}
\begin{aligned}
    \mathbf{C}^l=\{C_k \subset\calP | &k = 1, \dots N ;
    \\ 
    &l=\lambda(\mathbf{p}_i)=\lambda(\mathbf{p}_j),
    \ 
    \forall\mathbf{p}_i,\mathbf{p}_j\in C_k\}.
\end{aligned}
\end{equation}
We further enhance these semantic clusters by approximating each of them as Gaussian distributions:
\begin{equation}
    \bm{\mu}_k = \frac{1}{|{C}_k|}\sum\mathbf{p}_i,
    \bm{\Sigma}_k = \frac{1}{|{C}_k|}\sum(\mathbf{p}_i-\bm{\mu}_k)^{\top}(\mathbf{p}_i-\bm{\mu}_k).
\end{equation}

The query scan and reference semantic map can be represented by a set of semantic Gaussian distributions $\mathbb{C}_{\mathcal{A}}=\left\{\mathcal{A}_i^l\sim\mathcal{N}\left(\mathbf{a}_i, \mathbf{\Sigma}_{\mathcal{A}_i}\right)\right\}$ and $\mathbb{C}_{\mathcal{B}}=\left\{\mathcal{B}_j^l\sim\mathcal{N}\left(\mathbf{b}_j, \mathbf{\Sigma}_{\mathcal{B}_j}\right)\right\}$ respectively. These semantic Gaussian distributions will then act as primitives for establishing correspondences and later pose estimation. This representation is beneficial for correspondence generation as the covariance depicts the shape information of each semantic instance, which serves as another metric for similarity check. Such semantic lifting also structures the query scan as a two-layer scene graph where we have the semantic segmented points at the lower level and semantic instances as vertexes at the upper level. Each edge in the scene encodes the spatial relationship as well as the semantic topological information between two semantic instances.

\begin{figure}[t]
\vspace{5pt}
\centering
\includegraphics[width=8.5cm]{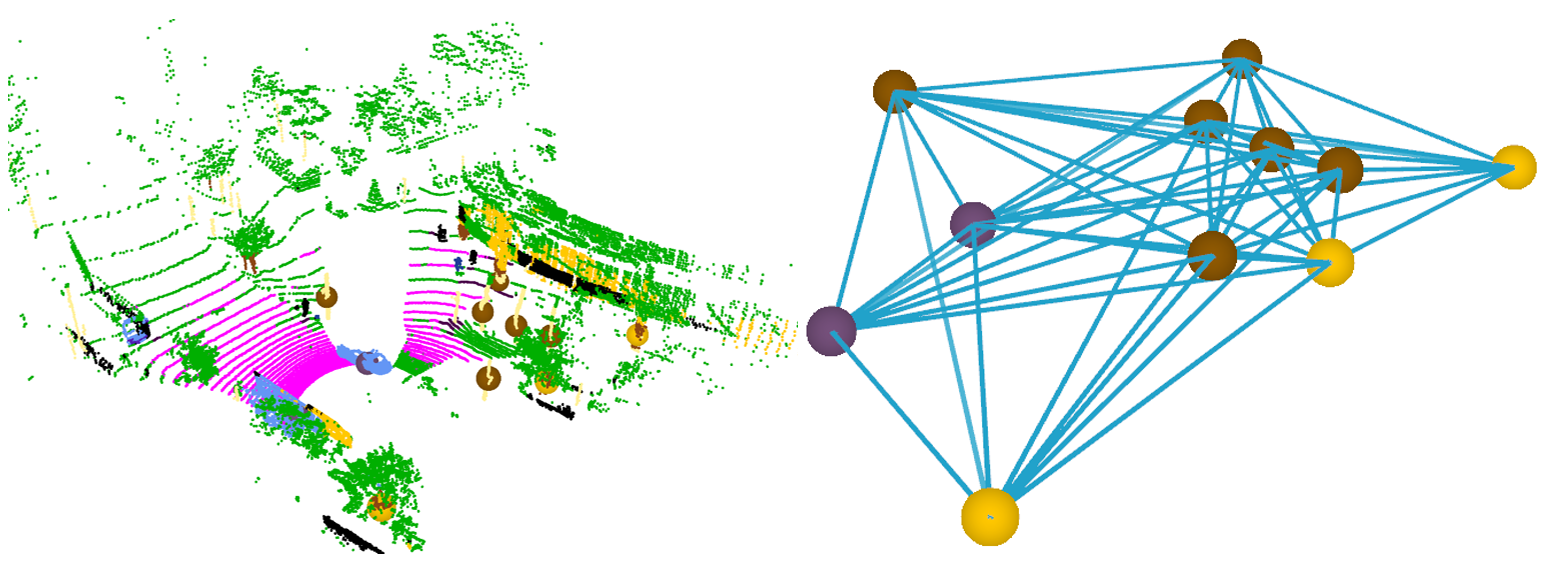}
\caption{One LiDAR scan from MulRan DCC dataset and its corresponding triangulated 3D scene graph. Colored spheres are the centroids of each semantic cluster with cars being purple, tree trunks being brown, and poles being yellow.}
\label{fig: tri scene graph}
\vspace{-1em}
\end{figure}

\subsection{Correspondence Generation via Local Substructures}

\begin{figure}[t]
\vspace{6pt}
\centering
\includegraphics[width=8.5cm]{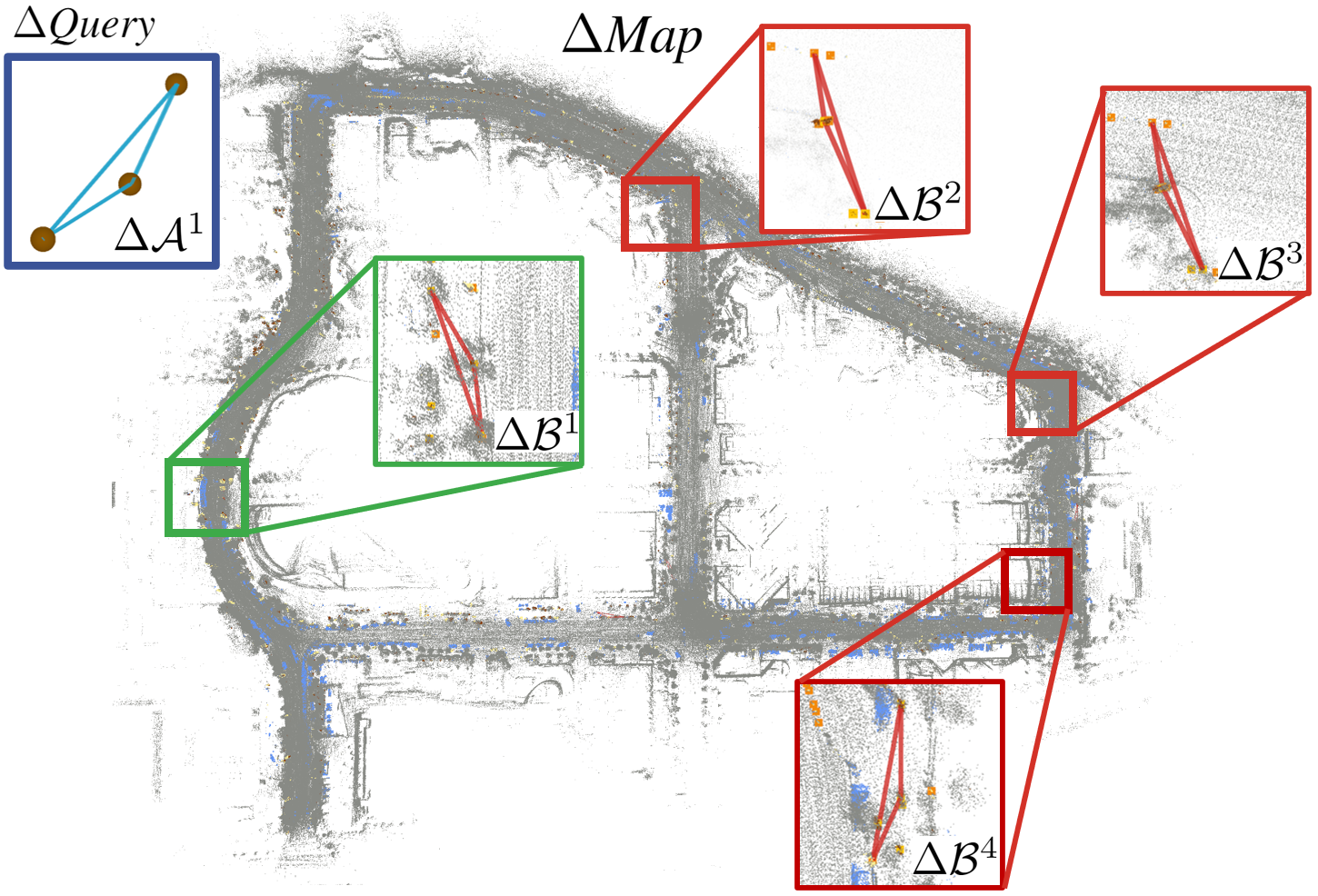}
\caption{Illustration of the substructure ambiguities and the proposed correspondence generation process. One triangle representation of the query scan (with three vertices labeled as tree trunk in brown) is shown in blue and multiple responses from different regions of the reference map are shown in green (true location) and red (false location) due to substructure ambiguities. Correspondence generation between all these substructures ensures local consistency while also retaining the possibility of exploiting scene-wise global consistency.}
\label{fig: corres gen}
\vspace{-1.3em}
\end{figure}

We then leverage substructures of the scene graph for correspondence generation. Inspired by STD \cite{yuan2023std}, we triangulate each scene graph to form a series of triangles as in the minimal representation for local similarity measurement and subsequent correspondence generation. To be more specific, each anchor semantic cluster $\mathcal{A}_i\sim\mathcal{N}\left(\mathbf{a}_i, \mathbf{\Sigma}_{\mathcal{A}_i}\right)$ is associated with $K$ nearest clusters $\left\{\mathcal{A}_j\right\}_{j=1}^K$. Afterward, we exhaustively select two of the neighbors, together with the anchor cluster, i.e., $\mathcal{A}_1$, $\mathcal{A}_2$ and $\mathcal{A}_3$, to form one triangle representation of current scene graph. By an abuse of notation, we denote it as $\Delta\left(\mathcal{A}_{1,2,3}\right)$ which comprises of the following attributes: 
\begin{itemize}
    \item $\mathbf{a}_1$, $\mathbf{a}_2$, $\mathbf{a}_3$: centroids of the semantic clusters;
    \item $\mathbf{\Sigma}_1$, $\mathbf{\Sigma}_2$, $\mathbf{\Sigma}_3$: corresponding covariance matrices;
    \item $d_{12}$, $d_{23}$, $d_{31}$: three side lengths, $d_{12} \leq d_{23} \leq d_{31}$;
    \item $l_1$, $l_2$, $l_3$: three semantic labels associated with each vertex of the triangle.
\end{itemize}
Similar to STD \cite{yuan2023std}, we build a hash table using only the sorted side length $d_{12}$, $d_{23}$, and $d_{31}$ as the key value due to its simplicity and permutation invariance. Other attributes are left for verification purposes. In the searching process, we have the triangulated scene graph in the query scan and reference map:
\begin{equation}
\begin{aligned}
    \Delta{\textit{Query}} =\left\{\Delta\left(\mathcal{A}_{1,2,3}^n\right)\right\}_{n = 1}^{N}, \\
    \Delta{\textit{Map}} =\left\{\Delta\left(\mathcal{B}_{1,2,3}^m\right)\right\}_{m=1}^{M},
\end{aligned} 
\end{equation}
where $n$ and $m$ are indexes for triangle descriptors in the query and map scene graph respectively. We drop the subscript and denote $\Delta\left(\mathcal{A}_{1,2,3}^n\right)$ as $\Delta\mathcal{A}^n$ for clarity. As shown in Fig. (\ref{fig: corres gen}), query each of the triangles (e.g., $\Delta\mathcal{A}^1$) against the hash table constructed by the reference semantic scene graph will produce multiple responses $\left\{\Delta\mathcal{B}^q\right\}_{q = 1}^{Q}$ as similar substructures could exist throughout the whole mapping region. We further leverage the semantic labels $l$, as well as the covariance matrix $\mathbf{\Sigma}$, associated with each vertex for another round of similarity-check for semantic and shape resemblance. For semantic labels, we simply employ the equality condition. For the covariance matrices, Wasserstein distance is applied for similarity measurement. After querying all triangles in the query scene graph, a set of raw cluster-wise correspondence $\mathcal{I}_\text{raw}$ can be naturally built as the sorted side length offers direct mapping between semantic clusters. 

Although the descriptor-based retrieval process presented above is very similar to the one in any ordinary LCD, we highlight here that we neither solve for the pose nor produce multiple candidates. Instead, we leverage these locally similar substructures, i.e., the triangulated scene graph, to build the coarse correspondence. This is different from both local feature-aggregation-based LCD methods \cite{zhong2009intrinsic,steder2011place} and global feature-based LCD methods \cite{kim2021scan, cattaneo2022lcdnet} where local features are only stacked for retrieval purpose in the former, and post-retrieval correspondence generation for the latter. We implicitly formulate the candidate selection process in the correspondence generation stage which guarantees the local similarity while also retaining the possibility of exploring scene-wise global similarity in the next stage.

\subsection{Global Graph-theoretic Outlier Pruning}
With the prebuilt correspondence set $\mathcal{I_\text{raw}}$ that associates subareas of the current scene with locally similar ones in the reference map, we seek to find an area that maximizes the number of mutually consistent correspondences as well as maintains the consistency between these local structures:
\begin{align}
\begin{split}
    &\mathop{\max}_{\mathcal{I}\subset\mathcal{I}_\text{raw}} \quad \left|\mathcal{I}\right| \\ 
    &\text{s.t.}\ \mathcal{D}\left(\mathcal{I}_i, \mathcal{I}_j\right) \leq\epsilon,\ \forall \mathcal{I}_i, \mathcal{I}_j\in \mathcal{I},
    \label{equ: max inlier}
\end{split}
\end{align}
where $\mathcal{D}$ is a metric consistency check indicating whether two correspondences are mutually consistent with each other and $\epsilon$ is the threshold. Namely, for two correspondences $\mathcal{I}_i$ and $\mathcal{I}_j$, with their corresponding semantic clusters $\mathcal{A}_i, \mathcal{B}_i$ and $\mathcal{A}_j, \mathcal{B}_j$, a consistency check is defined as
\begin{equation}
    \mathcal{D}\left(\mathcal{I}_i, \mathcal{I}_j\right) \triangleq \text{dist}\left(\mathcal{A}_{ij}, \mathcal{B}_{ij}\right),
\end{equation}
with $\mathcal{A}_{ij}:=\mathcal{A}_{i} - \mathcal{A}_{j}$ and $\mathcal{B}_{ij}:=\mathcal{B}_{i} - \mathcal{B}_{j}$ distribution differences between semantic clusters. It is worth noting that the similarity check $\mathcal{D}$ can vary. From the simplest Euclidean distance-based \cite{yang2020teaser,ankenbauer2023global} to distribution distance-based \cite{yin2023segregator}.

Problem \ref{equ: max inlier} can be solved by formulating the correspondence set to a consistency graph $\mathcal{G}=\left(\mathcal{V}, \mathcal{E}\right)$, where each vertex in $\mathcal{V}$ represent one correspondence and each edge in $\mathcal{E}$ represents two correspondences are consistent with each other evaluated by similarity check $\mathcal{D}$. Afterward, finding the inlier set is equivalent to searching for the maximum clique of the consistency graph. We invite interested readers to refer to \cite{yang2020teaser} for more details. Maximum clique is a classic combinatorial problem in graph theory and is NP-complete. We leverage the PMC library \cite{rossi2015parallel} to solve it. We will also show how current state-of-the-art registration-based global localization algorithms \cite{ankenbauer2023global} are brittle to the hardness of the maximum clique problem when the problem scales up even with a powerful parallel solver.

From a holistic view, such graph-theoretic outlier pruning strategy embeds globally consistent on top of the coarse correspondence set generated by locally consistent triangulated scene graphs. Such a local-to-global scheme resists constructing the hard association between the query scan and one single scan in the keyframe database, which is usually conducted by voting or similarity check. Instead, it first exploits local structures of one scene to generate correspondences associating similar substructures. Afterward, relationships between these local fragments are considered in search of a place in the reference that is globally consistent between the local substructures. The process presented here is very similar to feature re-ranking methods \cite{vidanapathirana2023spectral} for LCD where the computation of pose does not happen immediately after candidate retrieval but goes through another re-ranking process for frame-wise consistency verification. While our proposed method works on a more informative lower level, the substructures of scene graphs, thus have a better possibility of reaching global consistency.

\subsection{Pose Estimation}
With the estimated inlier set $\mathcal{I}$, we formulate the objective function in Eq. (\ref{eq: generative func}) into the following truncated least squares (TLS) form to resist potential outliers further\cite{yang2020graduated}
\begin{equation}
    \hat{\mathbf{R}},\hat{\mathbf{t}}=\mathop{\arg\min}_{{\bf R}\in SO(3), {\bf t}\in \mathbb{R}^3}\sum\limits_{ij\in\hat{\mathcal{I}}}\min\left(\left\|\mathbf{p}_i-\mathbf{R}\mathbf{q}_j-\mathbf{t}\right\|_2,c_{ij}\right),
    \label{eq: final obj}
\end{equation}
with $c_{ij}$ the truncation parameter. Eq. (\ref{eq: final obj}) is then solved by leveraging Black-Rangarajan duality \cite{black1996unification} and graduated non-convexity (GNC) \cite{yang2020graduated}.

\section{Experimental Results}
\label{sec: results}
In this section, we compare our proposed method with several state-of-the-art one-shot global localization methods. All mentioned algorithms are implemented in C++ and tested on a PC with Intel i9-13900 and 32Gb RAM.

\textbf{Experimental Setup.} We evaluate our proposed method, Outram, on six different sequences of two publicly available datasets: MulRan \cite{kim2020mulran} and MCD \cite{mcdviral2023}. To mimic a real global localization or relocalization scenario, different from a loop closure detection setting, we intentionally involve temporal diversity between the mapping or descriptor generation session and the localization session from days to months. For each mapping sequence, we concatenate semantically annotated scans \cite{cao2023multi} by the ground truth pose to generate the semantic segmented reference map for registration-based methods. Three representative semantic classes are used for all semantic-related methods: pole, tree trunk, and car. For LCD-based global localization methods, frames in the mapping sequences are encoded to form a database for retrieval using scans in localization sequences. Statistics of the benchmark datasets are presented in Table \ref{tab: dataset stats}. The criteria for choosing the mapping sequence is the sequence that has the most coverage of the target area. We also highlight the time differences between the mapping and localization sessions ranging from several days to months, making our setup suitable for benchmarking global localization algorithms.

\textbf{Baselines.} We involved a variety of state-of-the-art loop closure detection methods, STD \cite{yuan2023std}, Scan Context \cite{kim2021scan} and GosMatch \cite{zhu2020gosmatch}, as well as recently developed registration-based global localization algorithms \cite{ankenbauer2023global} to benchmark the performance of each of the method. STD also leverages substructures of a scene for loop closure detection in a voting manner and GosMatch leverages semantic clusters for global descriptor generation. As they have certain sub-modules that are the same as our proposed one, we involve them to confirm that our proposition, the local-to-global method, can exploit more structural information of a LiDAR scan and have a better chance to be localized in a one-shot manner. As most of the methods have an open-sourced implementation \cite{yuan2023std, zhu2020gosmatch, kim2021scan}, we directly use them for comparison. As to the method proposed by Ankenbauer et al. \cite{ankenbauer2023global}, since it also leverages semantic objects for global localization, we share the same semantic clusters for a fair comparison.

\begin{table}[t]
\vspace{5pt}
\caption{Details of Evaluation Datasets}
\vspace{-1em}
    \begin{center}
    \begin{tabular}{c|cc|c}
    \toprule
    Mapping/Loc. Sequence & Length & Scan Number & Time Diff. \\
    \midrule
    \textit{Mapping:} \\
    MulRan DCC 03   & 5.7km & 7479 & - \\
    MulRan KAIST 02 & 6.3km & 8941 & - \\
    MCD NTU 01      & 3.8km & 6023 & - \\
    \midrule
    \textit{Localization:}\\
    MulRan DCC 01   & 4.9km   & 5542 & 20 days\\
    MulRan DCC 02   & 5.2km   & 7561 & 1 month\\
    MulRan KAIST 01 & 6.3km   & 8226 & 2 months\\
    MulRan KAIST 03 & 6.4km   & 8629 & 10 days\\
    MCD NTU 02      & 0.64km  & 2288 & 2 hours\\
    MCD NTU 13      & 1.23km & 2337  & 2 days\\
    \bottomrule
    \end{tabular}
    \end{center}
    \label{tab: dataset stats}
    \vspace{-2.5em}
\end{table}

\textbf{Metric.} We employ the ordinary relative pose error (RPE) to evaluate the accuracy of estimated pose $\hat{\mathbf{T}}$ with respect to the ground truth $\mathbf{T}$:
\begin{equation*}
    e_{trans}=\sqrt{\Delta{x}^2+\Delta{y}^2+\Delta{z}^2},
\end{equation*}
\begin{equation*}
    e_{rot}=\arccos{\left(trace(\Delta{{\bf T}})/2-1\right)},
\end{equation*}
with $\Delta\mathbf{T}=\mathbf{\hat{T}}\cdot\mathbf{T}^{-1}$ the transformation difference, and $\Delta{x}$, $\Delta{y}$, $\Delta{z}$ the positional entries of $\Delta\mathbf{T}$. We regard global localization results with $e_{trans} < 5$ meters and $e_{rot} < 10$ degrees are valid, which is generally the convergence region for local registration methods \cite{pomerleau2013comparing} for subsequent refinement.

\begin{table*}[htb]
\vspace{5pt}
\caption{Global Localization Performance Comparison}
    \vspace{-1em}
    \begin{center}
        
    \begin{tabular}{clccccccccc}
    \toprule
    & & \multicolumn{6}{c}{Successful Global Localization Rate [\%] $\uparrow$} & \multirow{3}{*}{ATE [Meter] $\downarrow$} & \multirow{3}{*}{ARE [Deg] $\downarrow$} & \multirow{3}{*}{Time [ms] $\downarrow$} \\
    \cmidrule(lr){3-8}
    & Dataset & \multicolumn{2}{c}{MulRan DCC} & \multicolumn{2}{c}{MulRan KAIST} & \multicolumn{2}{c}{MCD NTU} \\
    \cmidrule(lr){3-4}
    \cmidrule(lr){5-6}
    \cmidrule(lr){7-8}
    & Localization Seq. & 01 & 02 & 01 & 03 & 02 & 13 \\
    \midrule
    \multirow{2}{*}{\rotatebox{90}{LCD}} & STD \cite{yuan2023std} & 17.57 & 18.06 & \underline{49.61} & 38.96 & 66.64 & 34.27 & \textbf{0.23} & \textbf{0.54} & \textbf{7.18}\\
    & GOSMatch \cite{zhu2020gosmatch} & 48.61 & 50.17 & 35.93 & \underline{51.98} & 55.01 & 42.53 & 1.92 & 2.06 & 12.08\\
    \midrule
    \multirow{4}{*}{\rotatebox{90}{Regis.}} & Ankenbauer et al. (Original) \cite{ankenbauer2023global} & - & - & - & - & 77.92 & 82.54 & 0.47 & 2.03 & 1708.3\\
    & Ankenbauer et al. (Cons.) \cite{ankenbauer2023global} & 0.072 & 0.032 & 0.025 & 0.012 & - & - & 2.81 & 3.17 & 345.3\\
    & Outram Pure Geo. (\textbf{Ours}) & \underline{50.65} & \underline{66.93} & 41.90 & 42.17 & \underline{82.11} & \underline{82.76} & 0.69 & 2.70 & 323.8\\
    & Outram (\textbf{Ours}) & \textbf{82.53} & \textbf{90.48} & \textbf{84.41} & \textbf{85.64} & \textbf{99.65} & \textbf{95.42} & 0.40 & 1.83 & 306.8\\
    \bottomrule
    \end{tabular}
    \end{center}
    \label{tab: preform com}
    \vspace{-2.5em}
\end{table*}

\textbf{Results.} We present the experimental results in four aspects, including successful global localization rate, error distribution analysis, runtime analysis, and storage analysis.

\subsection{Successful Rate of Global Localization}
We present the results of LCD-based global localization methods on the upper side of Table \ref{tab: preform com} while the registration-based ones are on the lower side. Our proposed algorithm, Outram, outperforms all other methods by a margin.

We observed that without the proposed triangulated scene graph for correspondence generation, the method proposed by Ankenbauer et al. \cite{ankenbauer2023global} can hardly scale to bigger size problems as it generates correspondences between all semantic clusters in the current scan and the reference map that have the same label. For smaller-size data sets (e.g., the reference semantic map of NTU MCD only includes 1192 clusters), it performs relatively well as in this case, the straightforward all-to-all correspondence generation method guarantees the full inclusion of the inlier correspondences while being also computationally amenable. However, when the reference map scales to a bigger size (e.g., 5136 and 5103 semantic clusters in MulRan DCC and KAIST respectively), the original method quickly becomes computationally intractable, where we observe the algorithm drained all the 32Gb RAM of the test platform, making the program to crash. In such scenarios, we modified the method to a constrained version where we limited the size of semantic clusters in the query scan by random downsampling. However, the constrained algorithm performs poorly due to the failure of inlier inclusion. We also involve a pure geometric variant of Outram for the ablation study of semantic labels. In the implementation of the method, we disable the semantic attributes of each semantic cluster and produce a set of triangle descriptors with pure geometric information, similar to the representation proposed in STD \cite{yuan2023std}. Conversely, these pure geometric triangle descriptors are used to build up correspondence followed by a graph-theoretic outlier pruning process, like what is proposed in this paper. These comparisons demonstrate the effectiveness of the proposed triangulated scene graph in terms of more informative correspondence extraction (compared with existing registration-based methods) and the superior performance of the whole registration-based pipeline (compared with LCD-based methods), which verifies our claims in \ref{intro}.

\subsection{Error Distribution Analysis}
\begin{figure}[htb]
\centering
\subfigure[ECDF of translation error.]{\includegraphics[width=0.23\textwidth]{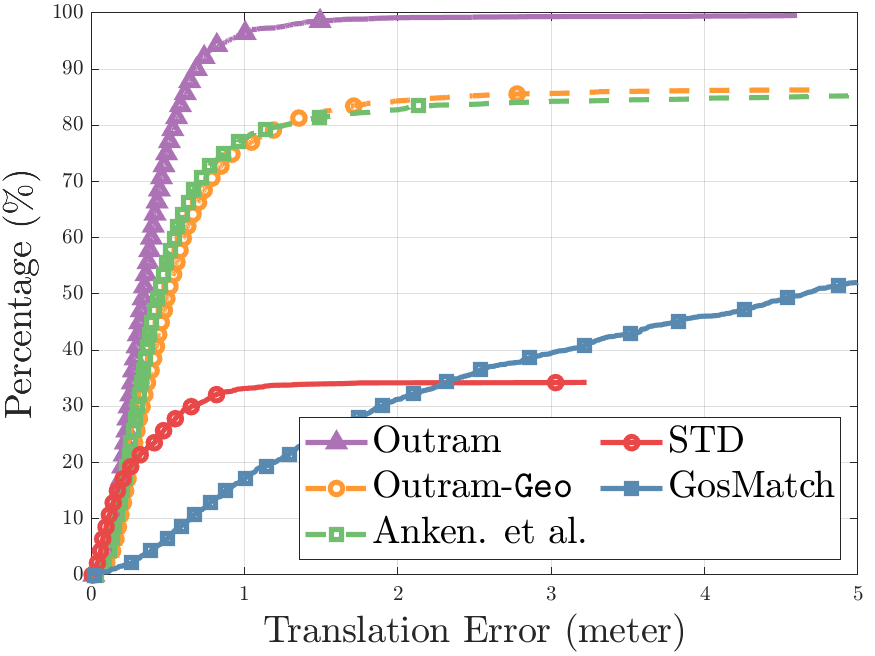} }
\subfigure[ECDF of rotation error.]{\includegraphics[width=0.23\textwidth]{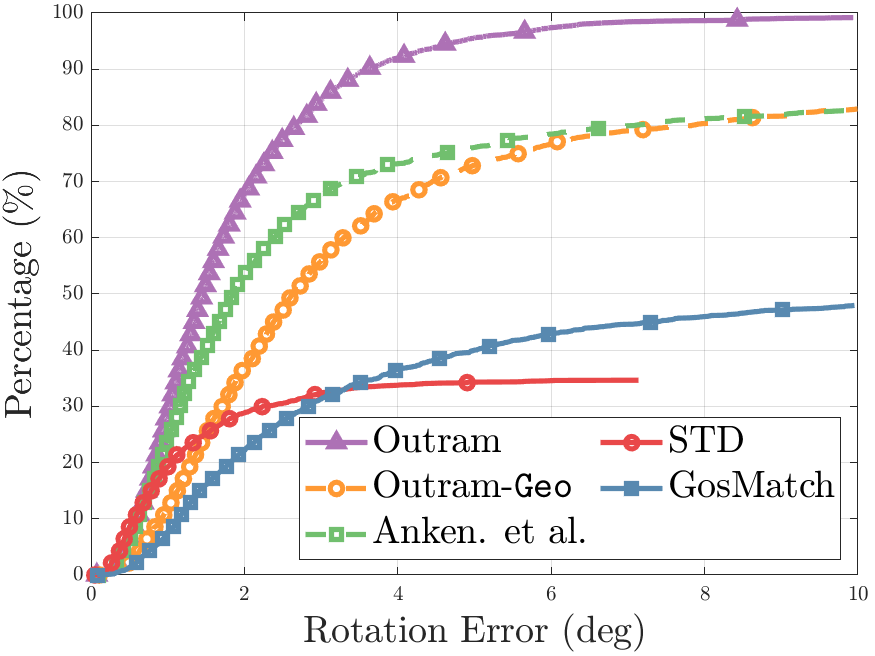} }
\caption{The empirical cumulative distribution function (ECDF) of translation and rotation error on sequence 13 of MCD NTU dataset. The x-axis represents different translation and rotation errors, and the corresponding y-axis is the probability of one specific method producing an estimation with a smaller error.}
\label{Fig: ECDF} 
\vspace{-2em}
\end{figure}

The average translation error (ATE) and average rotation error (ARE) in Table \ref{tab: preform com} are calculated using the successfully localized ones only. On average, point-based loop closure detection-based methods \cite{yuan2023std} have the most accurate localization result. The phenomenon can also be verified in Fig. (\ref{Fig: ECDF}) the cumulative distribution function (ECDF).
We observe that in the lower left corner, the state-of-the-art LCD-based global localization method, STD, surpasses all other methods in terms of translation error. This is because STD detects point-level stable features for pose estimation, while all other methods work on a higher level and rely on centroids of semantic clusters for pose estimation. Centroids could be different due to viewpoint variation thus affecting the pose estimation accuracy. 
However, this does not hinder the robustness of our proposed method to outperform others, which is the most important evaluation metric for global localization. 

\subsection{Runtime} 
In the last column of Table \ref{tab: preform com}, the average runtime of each method is presented. We noticed LCD-based methods are more computationally efficient compared with registration-based methods due to the simple retrieval-based design. All registration-based methods leverage maximum clique for outlier pruning, which could be time-consuming when the graph size is big.

\begin{figure}[t]
\centering
\includegraphics[width=8.5cm]{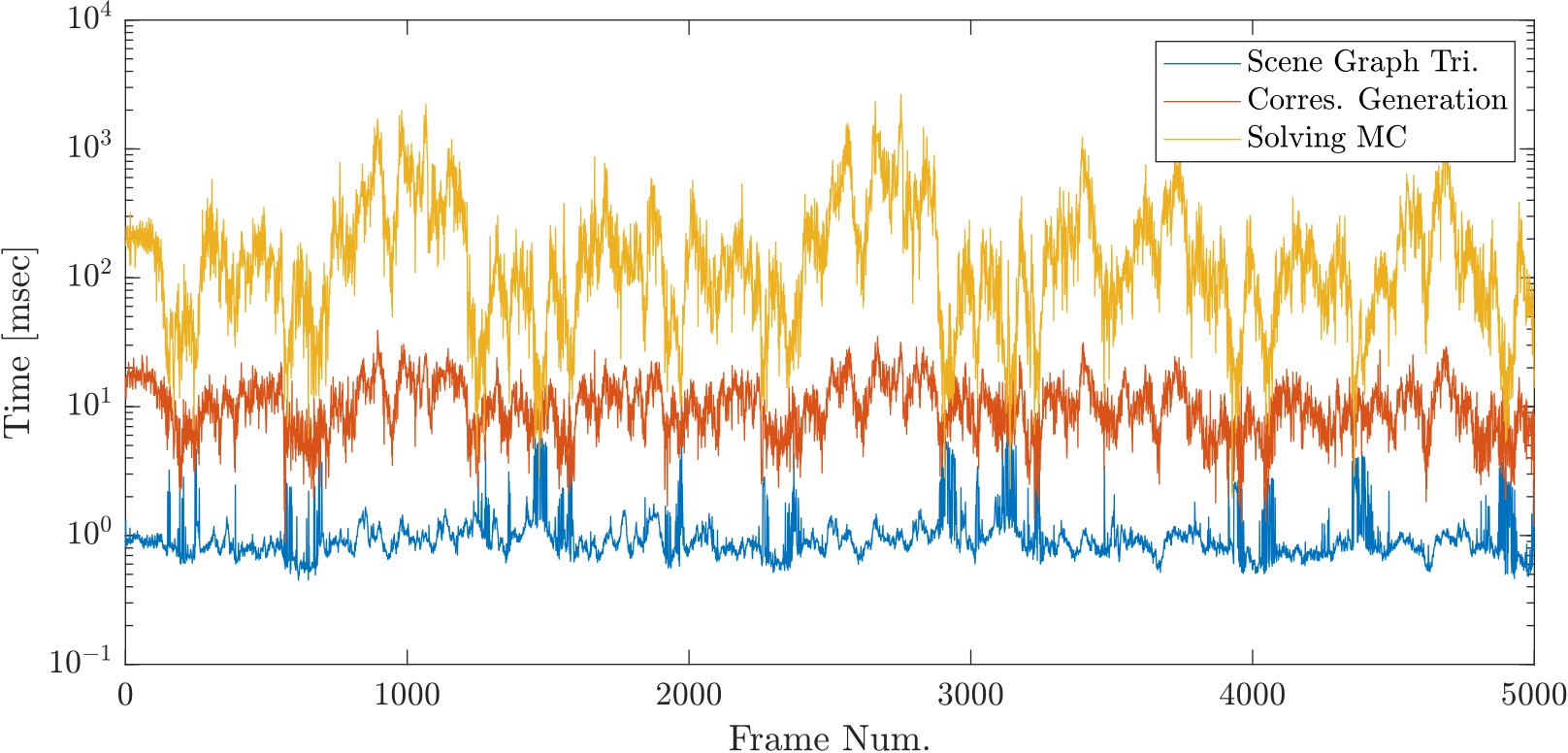}
\vspace{-5pt}
\caption{Runtime breakdown of Outram on MulRan DCC sequence 01.}
\label{fig: time breakdown}
\vspace{-0.2em}
\end{figure}

To understand each of the sub-modules of our proposed method better, we analyzed the time breakdown of each module and plotted it in Fig. (\ref{fig: time breakdown}). Three main components are considered here, namely, the time to generate a triangulated scene graph, the time to search for the corresponding substructure and establish correspondences, and the time to solve the subsequent maximum clique problem. Please note logarithmic scale is used in the y-axis for better visualization. We note that solving for the maximum clique (i.e., finding the globally consistent inlier correspondences out of the prebuilt one) requires 190 ms on average and is the most computationally expensive process. The time required for the process is jointly determined by the size of the 3D scene graph and the reference map. Although our system cannot run in real-time, the global localization task itself usually does not require a real-time algorithm.

\subsection{Storage Efficiency} 
\begin{table}[tb]
    \caption{Storage Consumption of Each Method on MulRan KAIST}
    \vspace{-10pt}
    \label{tab: storage comp}
    \begin{center}        
    \begin{tabular}{cccc}
    \toprule
    Method & STD & GosMatch & Regis.-based \\
    \midrule
    Storage Usage $\downarrow$ & 62.4MB & 4.8MB & \textbf{513kB} \\
    \bottomrule
    \end{tabular}
    \end{center}
    \vspace{-2.5em}
\end{table}

We further analyze the storage consumption of each method on MulRAN KAIST sequence 02 containing 8941 frames in Table \ref{tab: storage comp}. For average global descriptor-based LCD methods \cite{zhu2020gosmatch}, it requires several MB to store the vectorized descriptors for each frame. STD \cite{yuan2023std} requires storing planes detected in every frame for geometric verification, thus more space is consumed. The result reveals the potential of leveraging Outram for relocalization in bigger datasets.

\section{Conclusions}
In this paper, we propose Outram, a one-shot LiDAR global localization algorithm leveraging triangulated 3D scene graphs and graph-theoretic outlier pruning. Substructures of scene graphs are leveraged for locally consistent correspondence generation, and the subsequent outlier pruning process ensures the global consistency between the substructures and finds the inlier correspondences. We demonstrate the effectiveness of our proposed methods on various datasets where Outram surpasses several state-of-the-art LCD-based global localization methods, albeit at the cost of real-time performance. In the future, we plan to work on a proper indicator of the global localization quality and a theoretical guarantee for localization results.


{
\bibliographystyle{IEEEtran}
\bibliography{mybib}
}

\end{document}